%% file: main.tex
\documentclass[10pt,twocolumn,letterpaper]{article}

\usepackage{cvpr}              % To produce the CAMERA-READY version
\input{preamble}

\definecolor{cvprblue}{rgb}{0.21,0.49,0.74}
\usepackage[pagebackref,breaklinks,colorlinks,citecolor=cvprblue]{hyperref}

\def\paperID{11264} % *** Enter the Paper ID here
\def\confName{CVPR}
\def\confYear{2024}

\title{Part-aware Unified Representation of Language and Skeleton for Zero-shot Action Recognition}

\author{Anqi Zhu$^1$, Qiuhong Ke$^{2, }$\thanks{Corresponding author} ,  Mingming Gong$^1$, James Bailey$^1$\\
$^1$The University of Melbourne; $^2$Monash University\\
Parkville VIC 3052, Australia; Wellington Road, Clayton VIC 3800, Australia\\
{\tt\small azzh1@student.unimelb.edu.au, qiuhong.ke@monash.edu, \{mingming.gong, baileyj\}@unimelb.edu.au}}
\pgfplotsset{compat=1.18}
\begin{document}
\maketitle

\input{sec/0_abstract}    
\input{sec/1_intro}
\input{sec/2_related_works}

\input{sec/3_method}
\input{sec/4_experiments}
\input{sec/5_conclusion}
\input{sec/6_acknowledgement}
{
    \small
    \bibliographystyle{ieeenat_fullname}
    \bibliography{main}
}

% WARNING: do not forget to delete the supplementary pages from your submission 
% \input{sec/X_suppl}

\end{document}

%% file: preamble.tex
\usepackage[dvipsnames]{xcolor}

\usepackage{multirow}
\usepackage{stmaryrd}
\usepackage{tikz,pgfplots,wrapfig, bm}
\usepackage[flushleft]{threeparttable}
\usepackage{xpatch}
\usepackage{placeins}
\usepackage{tabularx}
\usepackage[accsupp]{axessibility} % Improves PDF readability for those with visual impairments.

\DeclareMathOperator*{\argmin}{arg\,min}

%% file: sec/0_abstract.tex
\begin{abstract}
%Skeleton-based action recognition is gaining popularity as a viable alternative to RGB-based classification due to its lower computational complexity and compact data structure. 
While remarkable progress has been made on supervised skeleton-based action recognition, the challenge of zero-shot recognition remains relatively unexplored. In this paper, we argue that relying solely on aligning label-level semantics and global skeleton features is insufficient to effectively transfer locally consistent visual knowledge from seen to unseen classes. To address this limitation, we introduce Part-aware Unified Representation between Language and Skeleton (PURLS) to explore visual-semantic alignment at both local and global scales. 
%\textcolor{red}{
PURLS introduces a new prompting module and a novel partitioning module to generate aligned textual and visual representations across different levels. The former leverages a pre-trained GPT-3 to infer refined descriptions of the %global/body-part-based local/temporal-interval-based-local movements 
global and local (body-part-based and temporal-interval-based) movements 
from the original action labels. % which are fed to and extracts their text embeddings using CLIP. 
The latter employs an adaptive sampling strategy to group visual features from all body joint movements that are semantically relevant to a given description. %During training, PURLS is trained to project an aligned visual-textual encoding manifold for each type of description in a balanced manner. 
Our approach is evaluated on various skeleton/language backbones
%}
and three large-scale datasets, \ie, \textit{NTU-RGB+D 60}, \textit{NTU-RGB+D 120}, and a newly curated dataset \textit{Kinetics-skeleton 200}. The results showcase the universality and superior performance of PURLS, surpassing prior skeleton-based solutions and standard baselines from other domains. The source codes can be accessed at \url{https://github.com/azzh1/PURLS}.
\end{abstract}

%% file: sec/1_intro.tex
\section{Introduction}
\label{sec:intro}
\begin{figure}[t]
  \centering
   \includegraphics[width=\linewidth]{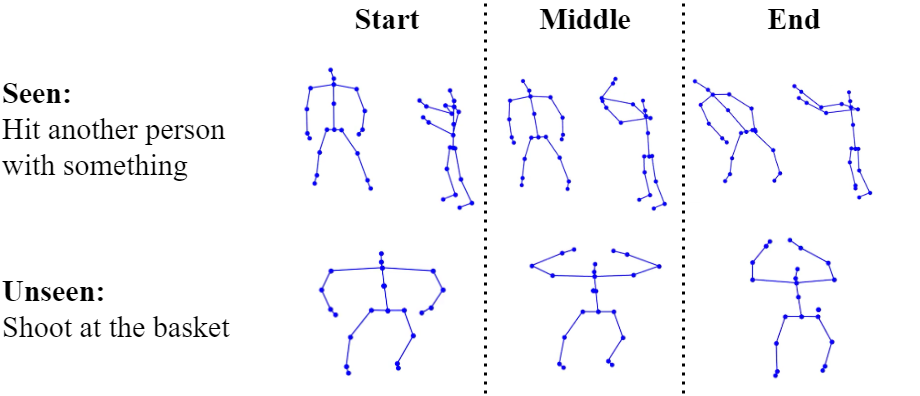}

   \caption{Examples of a seen class ('Hit another person with something') and an unseen class ('Shoot at the basket') from \textit{NTU-RGB+D 120} \cite{ntu120}. 
   % \textcolor{red}{
   While humans can quickly identify their similar hand movements and use this knowledge to distinguish the new class from other unseen classes, label-based global feature learning does not facilitate the direct transfer of such local knowledge.}
   % }
   \label{fig:zsl}
\end{figure}
Human action recognition (HAR) is an important topic in computer vision. As actions are the primary bridge for establishing communications between people and the outside world, HAR is used in many application domains, such as virtual reality \cite{vr1, vr2}, automated driving \cite{autodrive1, autodrive2}, video retrieval \cite{retrieval}, and robotics \cite{robotics1, robotics2}. The visual input modality can vary, covering RGB videos, depth image sequences, point clouds, and skeleton sequences \cite{har-survey}. During its early stages and even until today, the advancement of HAR has mainly been driven by RGB-based solutions due to their natural data abundance \cite{r2+1d, pyramid}. With the rise of pose prediction and depth-sensing technologies \cite{ntu60, ntu120, openpose}, 3-D skeleton sequences are becoming competitive substitutes that can reach high-accuracy prediction while cutting down computation, preserving profile privacy, and being robust by excluding background or color noises from action subjects. Due to these advantages, skeleton-based action recognition is attracting increasing attention in recent years \cite{app1, app2, app3}.

%Currently available skeleton datasets are primarily used for research on supervised action recognition. 
%Most existing research on skeleton-based action recognition \cite{deep-skel, deep-skel2, deep-skel3, 2s-gcn, dg-gcn, shift-gcn}  focuses on recognising actions in a fully supervised manner.
While remarkable progress has been made in this area, most existing research \cite{deep-skel, deep-skel2, deep-skel3, 2s-gcn, dg-gcn, shift-gcn} focuses on recognizing actions in a fully supervised manner, \ie, their designs require annotated data of all action classes for model training. Nevertheless, gathering labeled data for every potential action class is impractical, especially for rare or perilous actions.
Zero-shot learning (ZSL) is a research direction that aims to address this issue. 
% \textcolor{red}{
Previous ZSL approaches have focused on training models to align label embeddings with visual encoding outputs that are globally averaged from all skeleton features \cite{jpose, synse-zsl, smie}. However, as illustrated in \cref{fig:zsl}, actions that are globally dissimilar (\eg Hit another person with something vs. Shoot at the basket) may still exhibit similar local visual movements. The understanding of these shared movements should remain transferable across seen and unseen classes to enhance the prior knowledge for recognizing new actions. On the other hand, global semantic alignment concentrates on the cross-modal consistency of overall body actions and does not encapsulate refined learning on such local visual concepts. This constrains the generalization capability of the learned representation when applied to unseen classes.
% }
%semantically dissimilar actions (e.g. seen xx and unseen xx) can share similar local visual movements, and the knowledge of these movements should be transferrable between the classes to improve the recognition of the unseen actions. Defining an action as a kinetic combination of local motions conducted spatially and temporally, using global representations of the skeleton sequences is insufficient to    enable the textual semantic alignment  of such local visual concepts. This incapacity eventually restricts the generalization capability of the learned cross-modal representation on the unseen actions.

To overcome this issue, we present Part-aware Unified Representation of Language and Skeleton (PURLS), a novel framework that facilitates cross-modal semantic alignment at both global and local levels for prior knowledge exploitation.
% \textcolor{red}{
On the linguistic side, we start by enhancing the semantics of the original action labels using large language models.
Specifically, we design a prompting module that employs GPT-3 \cite{gpt3} to generate detailed descriptions for the original actions and their spatially/temporally divisible local movements (\ie, across human body parts/averaged temporal intervals). We then utilize a pre-trained text encoder from CLIP \cite{clip} to extract their textual features. For the visual aspect, a straightforward approach is to manually decompose a skeleton sequence into the corresponding global/local movement for a particular description and take the average features from the allocated body/temporal joints to perform alignment. Yet, this simple method limits the visual representation by strictly collecting features from static settings and thus may not always provide the most suitable alignment objects.
Hence, we introduce a unique partitioning module that adaptively finds the weights for each joint feature to correlate with the given descriptions. This eventually leads the model to provide a more semantically relevant visual representation for alignment. During training, PURLS learns to project the closest visual $\rightarrow$ textual manifolds at both global and local scales, ensuring semantic consistency with all descriptions in a balanced manner.
 This enables the projection layer to still distill part-aware knowledge when PURLS conducts prediction by only mapping global visual representations to label-level semantics during testing.
 % } %Therefore, when PURLS is pre-trained and used for classifying unseen classes, its prediction based on global feature matching outperforms traditional global-based methods. 

% \textcolor{red}{
Considering that PURLS is built upon feature-level operations, we test our model using multiple skeleton/language backbones, and compare the results with previous skeleton-based ZSL solutions and classic ZSL benchmarks from other domains. The experiments follow the existing evaluation setups on \textit{NTU-RGB+D 60} \cite{ntu60} and \textit{NTU-RGB+D 120} \cite{ntu120}, as well as additional setups with gradually increased unseen classes and decreased seen classes. We also evaluate the model's performance on a new dataset setting, \textit{Kinetics-skeleton 200}. Our results demonstrate that PURLS achieves state-of-the-art performance in all experiments and exhibits robust universality and generalizability. 
% } %, while other baselines experience significant accuracy drops in the new setups.

To summarize, our contributions are as follows:

\begin{itemize}

\item We propose  PURLS, a new framework for the exploration and alignment of global and local visual concepts with enriched semantic information for zero-shot action recognition on skeleton sequences. 

\item PURLS offers an adaptive weight learning approach for partitioning spatial/temporal local visual representations to support local knowledge transfer from seen to unseen actions.

\item PURLS is 
% \textcolor{red}{
steadily compatible with different feature extraction backbones, and
% }
achieves state-of-the-art performance on three of the public large-scale datasets for skeleton recognition.

\end{itemize}

%% file: sec/2_related_works.tex
\begin{figure*}[t]
  \centering
   \includegraphics[width=\textwidth]{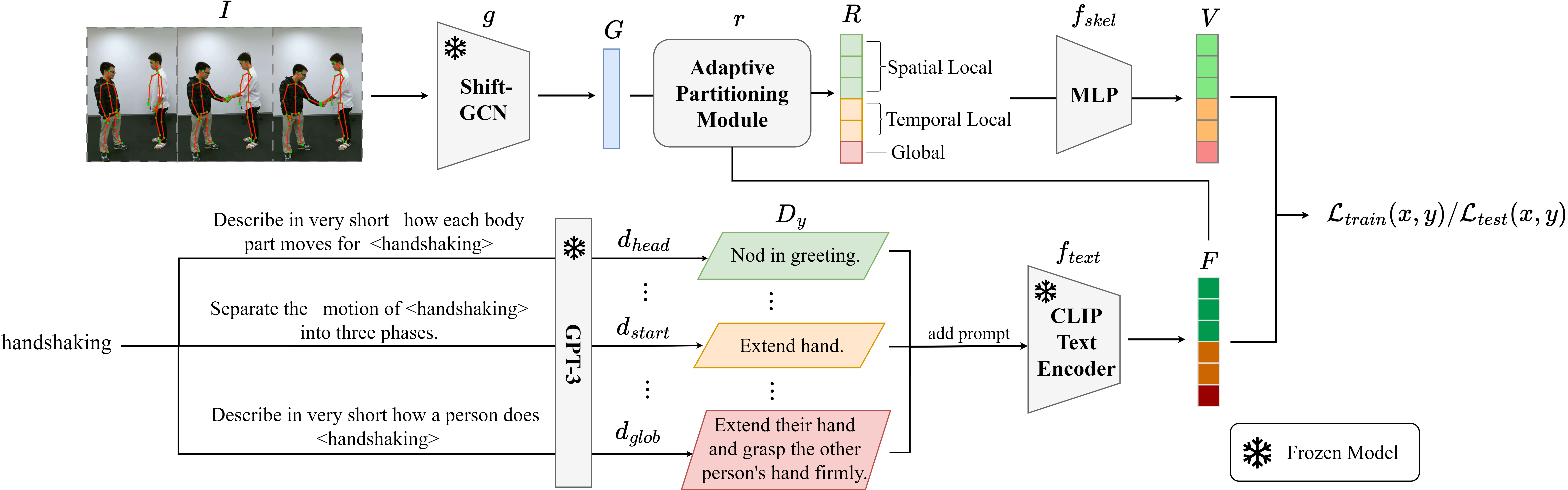}
  \caption{Architecture diagram for PURLS. 
  % \textcolor{red}{
  The matching action label is sent to GPT-3 \cite{gpt3} to obtain detailed descriptions for its global/local body movements, whose textual features are generated by a pre-trained language encoder of CLIP \cite{clip}. The visual features of the input skeleton sequence $I$ can be extracted from an arbitrary skeleton backbone $g$ (\eg, Shift-GCN \cite{shift-gcn}) pre-trained on the seen classes. The output $G$ is then fed to the partitioning module $r$ to group the joint-level features into global and spatially/temporally-local representations in an adaptive manner, which are later projected and aligned with their corresponding description embeddings.} 
  % }
  \label{fig:model-arc}
\end{figure*}

\section{Related Work}
\label{sec:literature}
%-----------------------------------------------------------
\subsection{Zero-shot Learning (ZSL)}
% \textcolor{red}{
Zero-shot Learning (ZSL) relies on training a model with samples from seen classes and their belonging class auxiliary information (\eg, text descriptions, pre-trained attribute features) to develop its recognition ability for unseen categories (assuming that their auxiliary information is also available). The goal of the training is to enable the model to establish a generalizable and meaningful connection between the new visual features and prior semantic knowledge for the unseen classes. The basic methods begin with embedding-based models \cite{emb1, emb2, emb3}, which directly construct a universal visual-to-textual projection to find the nearest label neighbors by cosine distances. Kodirov \cite{aezsl} pioneered using auto-encoders in ZSL, where the training target is to encode images into semantic space and then decode them back to visual signals. When recognizing unseen classes, the model can either use the pre-trained encoder to project image features to label semantics or decode language embeddings to visual dimensions as class prototypes. Butcher \cite{metric} enabled linear metric learning with a cross-modality alignment by creating a shared embedding space transformed from both visual and label encodings. From another perspective, mimicking human's learning habits, unseen subjects can be regarded as a new mix of visual concept components seen in training samples.
% }
\cite{semantic-disentangle} provided a generative approach in which the embedding alignment is established between disentangled local visual features and attribute-based text vectors. The disentanglement is realized by filtering out latent class-invariant features and verifying the decoding capacity of the remaining. In \cite{msdn}, a learnable attention module adaptively discovers the corresponding visual representation for each attribute. Applying similar intuition to RGB-based zero-shot action recognition, JigsawNet \cite{temp_align} recognized unseen actions by decomposing inputs in an unsupervised manner into atomic action prototypes that are pre-memorized from seen classes. 

% \textcolor{red}{
Compared to solutions that match global visual features and label semantics, predictions based on local representations or attributes usually achieve more precise and robust performance due to their wider knowledge transfer. However, these local-based solutions are developed for pixel-format inputs or disentangled global features, which are incompatible with the irregular graph-format feature of skeleton sequences. PURLS is the first paper to implement an automatic knowledge extraction and transfer for locally decomposable visual concepts hidden in skeleton kinematics.
% }

%-----------------------------------------------------------
\subsection{ZSL in Skeleton-based Action Recognition}
While diverse techniques have been developed for ZSL with RGB format inputs, limited research has been conducted on approaches against skeleton sequences. In 2019, \cite{jpose} proposed two standard methodologies adapted from traditional domains. These include a common-space metric learning using a Relation-Net framework and a visual $\rightarrow$ semantic embedding-based classification using a DeViSE \cite{devise} model. 
% \textcolor{red}{
Instead of learning to align every incoming pair of visual feature vectors and corresponding labels, \cite{smie} elucidated a more generalizable zero-shot prediction by learning to estimate and maximize the mutual information between overall visual and semantic distributions. In addition, the work designed a temporal rank loss to help the model capture more refined temporal information from frame-level visual features. SynSE-ZSL \cite{synse-zsl} was the first related work that considered skeleton-based local semantic matching. It learned a dual-modal feature representation by training the model to generate pseudo visual and linguistic samples from the opponent modality. The authors highlighted that action labels are often constituted by duplicated verb and noun phrases, and the visual patterns of most verbs are repetitively learnable from multiple seen classes. Therefore, discriminating the knowledge transfer of verbs from label-level semantics can effectively improve the model's generalization ability on unseen classes. We argue that a similar intuition can also be applied from the visual input side, in which PURLS mines spatially and temporally local visual concepts and uses language models to infer their aligned semantics from original labels.
% }

%-----------------------------------------------------------
\subsection{Multi-modal Representation Learning}
Multi-modal representation learning is highly relevant to ZSL, where different modalities can be mutually transformed and interpreted for information exchange. For the learning between image and language, CLIP\cite{clip} has provided a powerful backbone that utilizes contrastive learning to pre-train massive visual concepts from web data. The success of CLIP has constructed a universal representation manifold that captures shared semantics between RGB inputs and texts, which is widely used as a backbone reference for other downstream recognition tasks or ZSL baselines. In 3-D understanding, \cite{point-clip} enabled representation alignment for point cloud data by converting inputs into depth map images that fit the encoding format of CLIP. On the other hand, \cite{ulip} proposed ULIP, which directly unified the projected embeddings of images, texts, and point cloud values. Their experiments showed that distilling the knowledge from the matching language-image manifold can effectively overcome the generalization shortage in the original modality. 
% \textcolor{red}{
In skeleton learning, \cite{gap} explored importing comprehensive contrastive learning between static body-part-based skeleton features and their corresponding movement descriptions induced from original labels by GPT. Yet, their method relies on data-driven training and focuses on refining supervised recognition. In this paper, we provide PURLS to support adaptive knowledge transfer from seen to unseen classes according to a more powerful manifold alignment against local movements extractable either spatially or temporally.
% }

%% file: sec/3_method.tex
\begin{table*}[]
\centering
\resizebox{.9\textwidth}{!}{%
\begin{tabular}{|c|cccc|}
\hline
\multirow{2}{*}{\textbf{Action}} & \multicolumn{4}{c|}{\textbf{Question: Describe in   very short how each body part moves for \textless{}Action\textgreater{}.}}   \\ \cline{2-5} 
                                 & \multicolumn{1}{c|}{\textbf{Head}} & \multicolumn{1}{c|}{\textbf{Hands}} & \multicolumn{1}{c|}{\textbf{Torso}} & \textbf{Legs} \\ \hline
\begin{tabular}[c]{@{}c@{}}Hit another person \\      with something\end{tabular} &
  \multicolumn{1}{c|}{\begin{tabular}[c]{@{}c@{}}Turn towards the \\      other person.\end{tabular}} &
  \multicolumn{1}{c|}{\begin{tabular}[c]{@{}c@{}}Grip the object tightly \\
  and thrust it       forward.\end{tabular}} &
  \multicolumn{1}{c|}{\begin{tabular}[c]{@{}c@{}}Twist and turn to       generate \\      momentum for the strike.\end{tabular}} &
  \begin{tabular}[c]{@{}c@{}}Stomp the ground to provide \\      additional force for the strike.\end{tabular} \\ \hline
\begin{tabular}[c]{@{}c@{}}Shoot at the \\      basket\end{tabular} &
  \multicolumn{1}{c|}{\begin{tabular}[c]{@{}c@{}}Turn and look up \\      towards the basket.\end{tabular}} &
  \multicolumn{1}{c|}{\begin{tabular}[c]{@{}c@{}}Grip the basket\\       and release it.\end{tabular}} &
  \multicolumn{1}{c|}{\begin{tabular}[c]{@{}c@{}}Twist and extend      to generate \\power for the shot.\end{tabular}} &
  \begin{tabular}[c]{@{}c@{}}Bend slightly and \\      propel slightly upward.\end{tabular} \\ \hline
\end{tabular}%
}
\caption{Example body-part-based descriptions generated by GPT-3. The refined explanations correlate similar head and hand movements between `hit another person with something' and `shoot at the basket'. }
\label{table:gpt3}
\end{table*}

\begin{table*}[]
\centering
\resizebox{.85\textwidth}{!}{%
\begin{tabular}{|c|ccc|c|}
\hline
\multirow{2}{*}{\textbf{Action}} &
  \multicolumn{3}{c|}{\textbf{\begin{tabular}[c]{@{}c@{}}Question:   Separate the motion of \textless{}Action\textgreater{} into three phases.\end{tabular}}} &
  \multirow{2}{*}{\textbf{\begin{tabular}[c]{@{}c@{}}Question: Describe in very short \\ how a person does \textless{}Action\textgreater{}.\end{tabular}}} \\ \cline{2-4}
 &
  \multicolumn{1}{c|}{\textbf{Start}} &
  \multicolumn{1}{c|}{\textbf{Middle}} &
  \textbf{End} &
   \\ \hline
\begin{tabular}[c]{@{}c@{}}Hit another person \\ with something\end{tabular} &
  \multicolumn{1}{c|}{Raise   arm.} &
  \multicolumn{1}{c|}{Swing arm.} &
  Strike other person. &
  \begin{tabular}[c]{@{}c@{}}Swing their arm and strike the\\ other person with the object.\end{tabular} \\ \hline
\begin{tabular}[c]{@{}c@{}}Shoot at the \\ basket\end{tabular} &
  \multicolumn{1}{c|}{Raise   arm.} &
  \multicolumn{1}{c|}{Throw ball.} &
  Aim at basket. &
  \begin{tabular}[c]{@{}c@{}}Raise their arm and throw the ball \\ towards the basket.\end{tabular} \\ \hline
\end{tabular}%
}
\caption{Example temporal-interval-based and global descriptions generated by GPT-3. The refined explanations correlate similar starting global postures between `hit another person with something' and `shoot at the basket'. }
\label{table:gpt3-2}
\end{table*}

\section{Proposed Approach}
\label{methodology}
%The overall architecture of the proposed framework  PURLS is shown in  (\cref{fig:model-arc}). 
While distinct human actions may differ holistically, they often share similar local movements. The conventional training approach of directly aligning seen action labels with the overall representations of skeleton sequences fails to capture the semantic information of such local movements, thereby limiting the efficacy of zero-shot action recognition. To overcome this, PURLS adopts a two-step strategy. 
% \textcolor{red}{
As shown in \cref{fig:model-arc}, it first focuses on the nuanced descriptions of each action label, considering global, spatially local, and temporally local perspectives. Subsequently, it uses a unique adaptive partitioning module to generate and align the visual representations from the corresponding skeleton joint features with every derived description.
% }
In this section, we first list our problem definition and introduce how we generate descriptions for both global and local movements from the original action labels. Following that, we expound on the process of partitioning the skeleton sequences for optimal feature alignment.

%-----------------------------------------------------------
\subsection{Problem Definition}
Suppose $\mathcal{D}_{tr} = \{(x_{tr}^{sc}, y_{tr}^{sc})\}$ to be the set of $N_{tr}$ training samples from available seen classes $\mathcal{Y}^{sc}$. A skeleton sequence $x_{tr}^{sc}\in \mathbb{R}^{L\times J\times M\times 3}$ records the 3-D locations of $J$ body joints per actor in $L$ frames. $M$ is the maximum actor number per sequence. $y_{tr}^{sc}\in \mathcal{Y}^{sc}$ is the corresponding action label belonging to the seen class label set. Similarly, we let $\mathcal{D}_{te} = \{(x_{te}^{uc}, y_{te}^{uc})\}$ denote the set of $N_{te}$ testing samples from the unseen classes $\mathcal{Y}^{uc}$. Under a standard ZSL setting, we have $\mathcal{Y}^{sc} \cap \mathcal{Y}^{uc} = \phi$. Training with only seen class samples, we expect the model to learn an extensive alignment of feature representations between the visual and language modalities, whose knowledge is efficiently transferrable to predict $\hat{y}\in \mathcal{Y}^{uc}$ during evaluation.

%-----------------------------------------------------------------
\subsection{Creating Description-based Text Features}
\label{subsec: desc}
Inspired by human learning habits, we regard an action as a specific combination of local body movements that can be spatially or temporally decomposed. In addition to the label-level semantics, these local movements can also be independently learned  as individual visual concepts transferable across different classes. 
% \textcolor{red}{
To intelligently extract such underlying semantics, we adopt GPT-3 to produce textual descriptions for these movements at different scales.
% } 
\cref{table:gpt3} and \cref{table:gpt3-2} show the questions and example answers we used for generating local and global descriptions to enrich the original labels.  For local movements, we design to generate detailed descriptions that are individually performed either by $P$ ($P = 4$) body parts (\ie, `head', `hands', `torso', and `legs') or in $Z$ ($Z = 3$) contiguous temporal intervals (\eg, `start', `middle', and `end'). 
% \textcolor{red}{
To format the generated answers for each local part, we wrap the designed questions in a fixed prompt template
% }
as `Using the following format, \textless{}QUESTION\textgreater{}: \textless{}LOCAL PART $1$\textgreater{} would: ...; \textless{}LOCAL PART $2$\textgreater{} would ...; ...; \textless{LOCAL PART $H$\textgreater{} would: ....' where $H\in\{P, Z\}$ and \textless{}LOCAL PART $i$\textgreater{} refers to the corresponding local part name in $P$ body parts ($i\in [0, P)$) or $Z$ intervals ($i\in [0, Z)$). For the global semantic, we request GPT-3 to provide descriptions that augment the original label names. Note that one can also prepare different questions and generate multiple descriptions to calculate averaged text embeddings for later alignment. However, we consider that this does not lead to the key improvement in the later ZSL experiments, so we maintain using one question for each type of generation.
% \textcolor{red}{
After acquiring the targeted answers, we calculate their text embeddings using a pre-trained CLIP \cite{clip} text encoder $f_{text}$ after converting them into standard prompts as ``a (cropped/trimmed) video of [DESCRIPTION]". For a given original label of $y\in \mathcal{Y}^{sc} \cup \mathcal{Y}^{uc}$, after GPT-3 produces its relevant descriptions $D_y = \{d_{head}, d_{hands}, ..., d_{start}, ..., d_{end}, d_{glob}\}$, we have $F = \bigparallel_d^{D_y}f_{text}(d) \in \mathbb{R}^{(P+Z+1)\times m}$ in which $m$ refers to the text embedding dimension size and $(P+Z+1)$ denotes the concatenation of $P$ body-part-based, $Z$ interval-based, and one global-based semantic encodings. 
% }

%-----------------------------------------------------------------
\subsection{Partitioning Skeleton Feature Representations}
\label{subsec:partition}
Following \cite{synse-zsl}, we first conduct the same padding and normalization pre-process from \cite{shift-gcn} to get the standard input $I\in \mathbb{R}^{L\times J\times M\times 3}$ of a raw skeleton sequence $x$ and then adopt a pre-trained Shift-GCN \cite{shift-gcn} to extract its visual features $G = g(I)\in \mathbb{R}^{S\times n}$, where $S = L'\times J$. $n$ is the skeleton encoding dimension and $L'$ is the temporal feature dimension size after $I$ being convoluted in $g$. To simplify the calculation, we average the features for $M$ performers. 
% \textcolor{red}{
To align with the output from the language branch, a partitioning module is further applied to extract the corresponding local and global representations from $G$.
% }  %which are eventually concatenated as a vector $R = r(G)\in\mathbb{R}^{(P+Z+1)\times m}$. Here, we provide two grouping principles $r$ relying on either pre-defined or adaptive partitioning schemes.  

 A straightforward method to generate spatially local representations involves manually breaking down the skeleton joints into body parts as shown in \cref{fig:pt}.  The feature of each body part  can then be derived by averaging the features of the joints inside itself over the whole sequence.
 For temporal partitioning, one can averagely divide $G$ along temporal  
dimensions into $Z$ consecutive segments. The representation for each segment  can then be computed by averaging the features of all body joints over the segment.
The global representation  can be achieved by averaging the features of all body joints over the whole sequence. 
We refer to this method as \textbf{static partitioning}. 
% \textcolor{red}{
While simple, these pre-defined partitions often exhibit instability in presenting the visual information that matches their corresponding descriptions. Below, we discuss our reasoning behind this observation and present our solution to resolve the issue.
% }

%Eventually, all the representations can be concatenated to  get $R = \bigparallel\limits_{p=1}^P R_{B_p} \bigparallel\limits_{z=1}^Z R_z \concat R_{glob} \in \mathbb{R}^{(P+Z+1)\times n}$ as the visual output ready to be transformed for semantic alignment.

\begin{figure}[t]
  \centering
   \includegraphics[width=.5\linewidth]{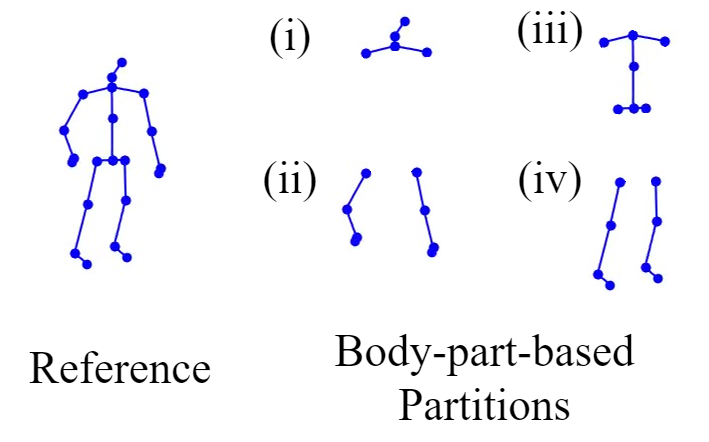}

   \caption{Spatial partitioning scheme for decomposing  body joints into four body parts: (i) Head, (ii) Hands, (iii) Torso, (iv) Legs.  }
   \label{fig:pt}
\end{figure}

\begin{figure}[t]
  \centering
   \includegraphics[width=\linewidth]{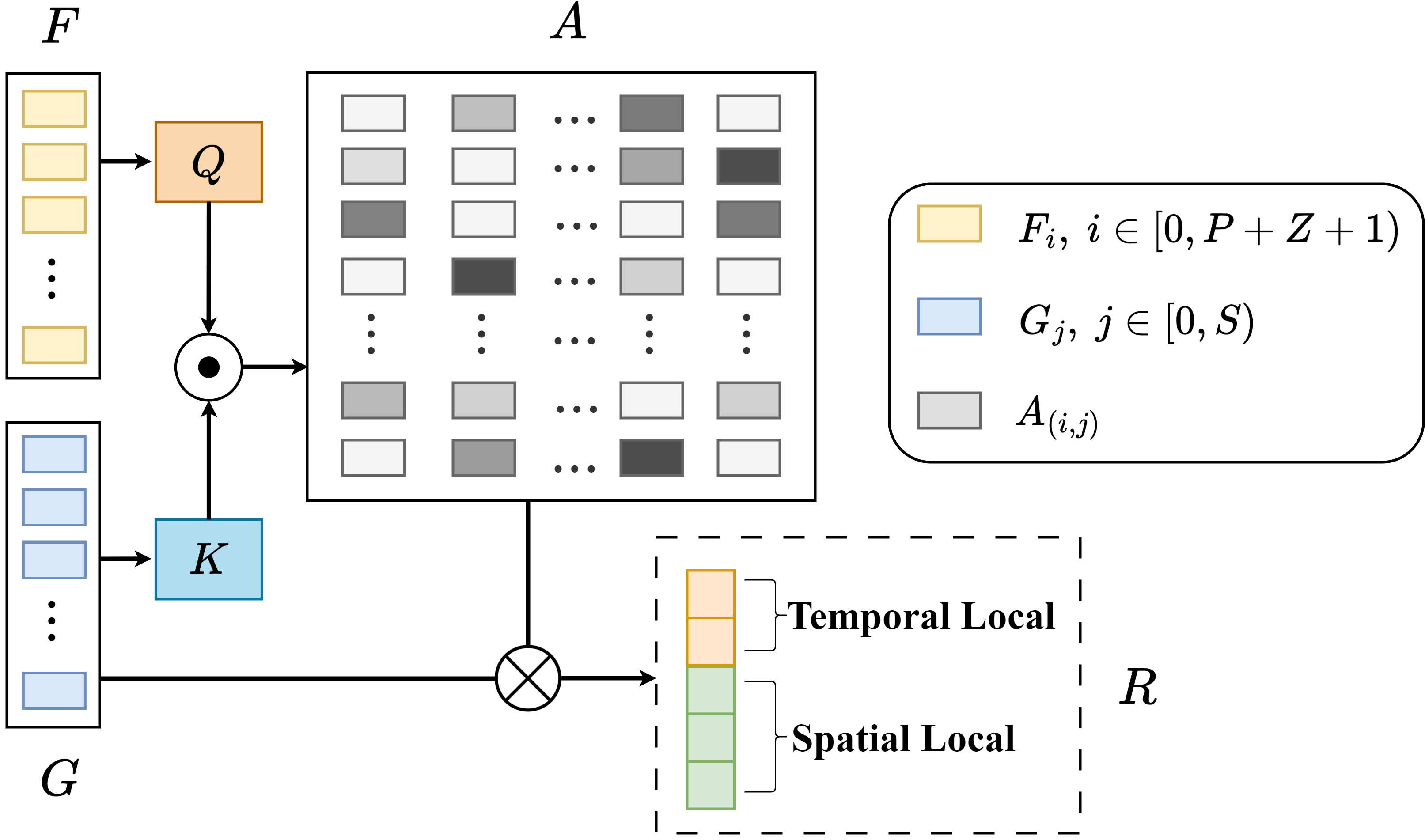}

   \caption{Illustration of how the adaptive partitioning module samples  local visual representations. %Temporal-based representations are produced in a similar way.
   }
   \label{fig:adaptive}
\end{figure}

\textbf{Adaptive partitioning:} 
% \textcolor{red}{
Static partitioning extracts local movements from fixed allocated node features. This requires considerable manual examination of potential training datasets to determine the most suitable division principles. Furthermore, local recognition can benefit from detecting its combinative context postures in other body parts or intervals. For example, the leg-lifting movement for walking can be more robustly recognized by simultaneously acknowledging an arm-swinging movement. Therefore, a more flexible approach for generating a local representation is to adaptively sample all description-relevant node features from $G$. \cref{fig:adaptive} presents our cross-attention-based adaptive partitioning module. To represent a particular body part or interval, we identify the nodes semantically related to its description in terms of spatial and temporal dimensions and then contribute their visual information based on a correlation weight. Specifically, with the text embedding $F$ and the visual output $G$, 
we define $Q = FW_Q$ as the language queries, and $K = GW_K$ as the visual keys. $W_Q\in \mathbb{R}^{m\times h}$ and $W_K \in \mathbb{R}^{n\times h}$ are learnable linear transformation matrices, where $h$ is the projection dimension size. %with a size of $m\times a$ and $n\times a$. %whose sizes are $m\times h$ and $n\times h$. 
The module estimates an attention matrix $A\in \mathbb{R}^{(P+Z+1)\times S}$ by applying a cross product between $Q$ and $K$, followed by a normalization and a softmax process as
% }
\begin{equation}
A = softmax((Q\times K^T) / \sqrt{h}). 
\end{equation}
% \textcolor{red}{
Intuitively, denoting the  $i$-th row of $A$ as $A_i\in \mathbb{R}^{1\times S}$, it calculates the respective semantic relevance for all $S$ nodes against the $i$-th description $d_i$ ($i\in[0, P+Z+1)$). Hence, the paired visual representation  $R_i\in\mathbb{R}^n$ for $d_i$ can be computed as a weighted sum from all node features, where the weights are defined by $A_i$. Promoting this to a matrix-level calculation, we can compute the paired visual representations $R\in \mathbb{R}^{(P+Z+1)\times n}$ for all $P+Z+1$ descriptions as
% }
\begin{equation}
  R = A G.
  \label{eq:attp}
\end{equation}

%where $i\in[0, P+Z]$. After obtaining the visual vector for each local/global description, we concatenate each $R_i$ to obtain the integrated visual representation $R\in\mathbb{R}^{(P+Z+1)\times n}$ ready for being transformed into text embeddings.
%\begin{equation}
 % R = \bigparallel\nolimits_{i=1}^{P+Z+1} R_i 
%\end{equation}

%--------------------------------------------------------
% \begin{MyColorPar}{red}
\subsection{Aligning dual-modal representations}
A pre-trained CLIP model \cite{clip} understands a wide range of visual concepts shared among natural language and images. To realize part-aware matching between skeleton sequences and texts, we design PURLS to evenly map each visual representation $R_i \in \mathbb{R}^n$ (the $i$-th row of $R$) to an aligned distribution of its description encoding $F_i\in \mathbb{R}^m$ (the $i$-th row of $F$). As shown in \cref{fig:model-arc}, we construct an MLP layer $f_{skel}$ to project each $R_i$ to the textual embedding space as $V_i\in\mathbb{R}^{m}$. Then, we conduct contrastive learning between ${V}_i$ and $F_i$ as follows:
\begin{equation}
\begin{aligned}
\mathcal{L}(V_i, F_i) = &-\frac{1}{2}\log\frac{\exp(\frac{V_i F_i}{\tau})}{\sum_{o}^{\mathcal{Y}^{sc}}\exp(\frac{V_i F_i^o}{\tau})} \\
  &-\frac{1}{2}\log\frac{\exp(\frac{V_i F_i}{\tau})}{\sum_w\exp(\frac{V_i^w F_i}{\tau}),}
\end{aligned}
  \label{eq:cl}
\end{equation}
where $F_i^o$ refers to the text embedding of the $i$-th description from other $o$ (negative) seen action classes, and $V_i^w$ is the $i$-th projected visual embedding from other $w$ (negative) skeleton samples in the same batch. The temperature parameter $\tau$ controls the training gradient.
%With Eqn\ref{eq:cl}, we can obtain the alignment losses for the local $P$ body parts $\mathcal{L}_{locP}$, local $Z$ temporal segments $\mathcal{L}_{locZ}$ and the global
  The overall training loss of PURLS is a weighted sum of alignment losses for all local and global representations:
\begin{equation}
\mathcal{L}_{train} (x, y) = \sum\nolimits_{i=0}^{P+Z}\alpha_i \mathcal{L}(V_i, F_i),
\label{eq:loss}
\end{equation}
in which the weights  $\alpha_i$  are either set to $1 / (P+Z+1)$ or learnable. %,  to distribute the importance of learning the $i$-th representation for final prediction.
During the training, $f_{skel}$ controls the visual$\rightarrow$ textual mapping for all representations and thus learns to adaptively balance the semantic alignments for local and global projections. This helps $f_{skel}$ to distill local-aware interaction knowledge when it encodes for global representations. Therefore, at the testing stage, given an input $x_{te}$, we can simplify the inference and predict $\hat{y}_{te}$ that yields the lowest alignment loss directly on its global representation:
\begin{equation}
\mathcal{L}_{test}(x_{te}, y) = -\frac{1}{2}\log\frac{\exp(\frac{V_k^{te} F_k}{\tau})}{\sum_{o}^{\mathcal{Y}^{uc}}\exp(\frac{V_k^{te} F_k^o}{\tau})},\ k = P + Z,
\end{equation}
\begin{equation}
\hat{y}_{te} = \argmin_{y\in\mathcal{Y}^{uc}} \mathcal{L}_{test}(x_{te}, y).
\end{equation}
% We only use the global prediction as it is more computationally efficient, and the mapping function has already encapsulated the relevant knowledge regarding local semantic interactions during the training process.
% \end{MyColorPar}

%% file: sec/4_experiments.tex
\section{Experiments}
\subsection{Datasets}
\label{subsec: datasets}
\textbf{\textit{NTU-RGB+D 60}} \cite{ntu60} contains $56,880$ skeleton sequence samples of $60$ actions, with $40$ individual subjects captured from $80$ distinct camera viewpoints. Each sample provides a temporal sequence of the 3-D location coordinates for $25$ human body joints per performer. The maximum performer number is $2$, and the coordinate values are padded as $0$ when the corresponding performer is unavailable (\eg single-person actions). We use the two splits suggested by \cite{synse-zsl} - a $55/5$ split (with $55$ seen classes and $5$ randomly chosen unseen classes) and a $48/12$ split. We then create two more difficult splits of $40/20$ and $30/30$ to further challenge the generalization ability of our solutions on more unseen classes with less available training.  

\textbf{\textit{NTU-RGB+D 120}} \cite{ntu120} is an enlarged dataset based on \textit{NTU-RGB+D 60} and includes $60$ additional action classes. It contains $114,480$ samples for $120$ actions performed by $106$ individual subjects captured from $155$ distinct camera viewpoints. Analogous to the above, we use two existing splits of $110/10$ and $96/24$ in \cite{synse-zsl} and two new splits of $80/40$ and $60/60$ for our evaluation setups.

\textbf{\textit{Kinetics-skeleton 200}} is a customized subset containing samples from the first $200$ classes of the \textit{Kinetics-skeleton 400} dataset \cite{st-gcn}. 
% \textcolor{red}{
When reviewing the ZSL experimental setups in other domains, we find that the existing protocols for skeleton understanding are very limited, as only the \textit{NTU-RGB+D} series provides standard ZSL benchmarks. This motivates us to establish initial benchmarks on other common action datasets. \textit{Kinetics-skeleton 400} includes the skeleton sequences extracted from the samples of $400$ human action classes in \textit{Kinetics 400} \cite{kinetic}. Its classes range from daily activities to complex actions. Each category contains at least $400$ YouTube video clips, from which the skeleton data is extracted using OpenPose \cite{openpose} in \cite{st-gcn}. During the experiments, we observed that the ZSL accuracy gradually diminished as the number of unseen classes increased to a certain large number. This is probably because the difficulty of the task eventually surpasses the feature extraction capacity of the pre-trained visual backbones, which is not a research focus of our paper. Therefore, we limit our learning scenarios to cover under $200$ classes. Similarly, we create four splits of $180/20$, $160/40$, $140/60$, and $120/80$ for our setups. For the full experiment results on the complete \textit{Kinetics-skeleton 400} and other small datasets (\textit{NW-UCLA} \cite{nwucla}, \textit{UTD-MHAD} \cite{utdmhad}, and \textit{UWA3D II} \cite{uwa3dii}), please refer to our supplementary materials.
% }
\begin{table*}[]
\centering
\resizebox{.8\textwidth}{!}{%
\begin{threeparttable}
\begin{tabular}{ccccccccccccc}
\toprule
\multirow{2}{*}{\textbf{Model}} &
  \multicolumn{4}{c}{\textbf{NTU-RGBD 60 (Acc \%)}} &
  \multicolumn{4}{c}{\textbf{NTU-RGBD 120 (Acc \%)}} &
  \multicolumn{4}{c}{\textbf{Kinetics-skeleton 200 (Acc \%)}} \\ 
        & 55/5           & 48/12 & 40/20 & 30/30 & 110/10 & 96/24 & 80/40 & 60/60 & 180/20 & 160/40 & 140/60        & 120/80 \\ \midrule
ReViSE \cite{revise}        & 75.37          & 26.44 & 24.26 & 14.81 & 57.92  & 37.96 & 19.47 & 8.27  & 24.95  & 13.28  & 8.14          & 6.23   \\ 
DeViSE \cite{devise}        & 77.61          & 35.80  & 26.91 & 18.45 & 61.52  & 40.91 & 19.50  & 12.19 & 22.22  & 12.32  & 7.97          & 5.65   \\ 
JPoSE \cite{jpose} & 64.82          & 28.75 & 20.05     & 12.39     & 51.93  & 32.44 & 13.71     & 7.65     & -      & -      & -             & -      \\ 
CADA-VAE \cite{cada-vae}      & 76.84          & 28.96 & 16.21     & 11.51     & 59.53  & 35.77 & 10.55     & 5.67     & -      & -      & -             & -      \\ 
SynSE \cite{synse-zsl}         & 75.81          & 33.30  & 19.85     & 12.00     & 62.69  & 38.70  & 13.64     & 7.73     & -      & -      & -             & -      \\ 
SMIE \cite{smie}      & 77.98          & 40.18 & -     & -     & 65.74  & 45.30 & -     & -     & -      & -      & -             & -      \\ 
    Global      & 64.69 & 35.46 & 27.15 & 16.29 & 66.96  & 44.27 & 21.31 & 14.12 & 25.96  & 15.85   & 10.23          & 7.77   \\
    \midrule
PURLS   &
  \textbf{79.23} &
  \textbf{40.99} &
  \textbf{31.05} &
  \textbf{23.52} &
  \textbf{71.95} &
  \textbf{52.01} &
  \textbf{28.38} &
  \textbf{19.63} &
  \textbf{32.22} &
  \textbf{22.56} &
  \textbf{12.01} &
  \textbf{11.75} \\ \bottomrule
\end{tabular}%
  \end{threeparttable}
}
\caption{Zero-shot action recognition results (\%) on \textit{NTU-RGB+D 60}, \textit{NTU-RGB+D 120} and \textit{Kinetics-skeleton 200}. Experiments for JPoSE, CADA-VAE, and SynSE on \textit{Kinetic-skeleton 200} are omitted because their pre-trained text features from their work are not consistent with other customized approaches. The experiments for SMIE is excerpted from its original paper.}
\label{table:results}
\end{table*}
\subsection{Implementation Details}
\label{subsec: implementation}
To prepare the skeleton backbone, we follow \cite{synse-zsl} and only use seen class samples to pre-train feature extraction for the setup of each split. The visual features are realized by the 256-dimensional penultimate layer feature from Shift-GCN \cite{shift-gcn} ($n=256$). We use the GPT-3 DaVinci-003 model and the questions in \cref{table:gpt3} and \cref{table:gpt3-2} to generate detailed descriptions for the original action labels. The textual features are then realized by the 512-dimensional encoding output from CLIP \cite{clip} equipped with the frozen weight of ViT-B/32 ($m=512$). For the architectural details of PURLS, we always set $P = 4$, $Z = 3$, %When using static partitioning, the model only contains a 2-layer MLP that projects the $(P+Z+1)\times 256$-dim visual features to $(P+Z+1)\times 512$-dim textual embeddings. For adaptive partitioning, %the model has an extra attention module in which 
$W_Q \in \mathbb{R}^{512\times 150}, W_K \in \mathbb{R}^{256\times 150}$ where $h = 150$. $f_{skel}$ is a 2-layer MLP in which the size of each hidden layer is 512. In the experiments on \textit{NTU-RGB+D 60} and \textit{NTU-RGB+D 120}, we set $L = 300$, $J = 25$, $M = 2$. For \textit{Kinetic-skeletons 200}, we adopt the same data input configs from \cite{st-gcn}, where $L = 300$, $J = 18$, $M = 2$. 

For training details, the model is optimized by an Adam optimizer with a learning rate of $1e-4$ and a batch size of 256. The training epoch number is set to 300 but allows early stops if the training accuracy does not improve in the latest 20 epochs. All of our experiments are conducted using PyTorch on one A100 GPU.

For experiment details, since few previous works are available for skeleton-based ZSL, we implemented some classic baselines used in RGB-based classification from scratch and also referred to the existing skeleton ZSL solutions from \cite{synse-zsl} and \cite{smie}. These include visual-to-language embedding models (DeViSE \cite{devise}, JPoSE \cite{jpose}), common-space embedding models (ReViSE \cite{revise}), generative solutions (CADA-VAE \cite{cada-vae}, SynSE \cite{synse-zsl}) and contrastive learning (SMIE \cite{smie}). Additionally, we have another baseline that only learns from the global feature alignment with the label-level encoding from CLIP. We mark this method as `\textbf{Global}' in all of our evaluation tables. While JPoSE, CADA-VAE, SynSE, and SMIE have their original linguistic feature configurations, the text features used in other customized baselines are uniformly encoded by the same CLIP we use for PURLS. The results for ReViSE and DeViSE are better than their records in the previous papers \cite{smie, synse-zsl} as they use better language models for text embedding.

%--------------------------------------------------------

\subsection{Results \& Analysis}
\label{subsec: results}
% results

\cref{table:results} presents the classification results using all mentioned baselines and PURLS under the given setups. The learning difficulty increases in the order of \textit{NTU-RGB+D 60}, \textit{NTU-RGB+D 120}, and \textit{Kinetics-skeleton 200}. Under the same dataset, the setups become more challenging with the decrease of seen classes and the increase of unseen classes. 

% \textcolor{red}{
Our method gives the highest performing predictions in every experimental setting. We observe that all previous baselines experience different levels of generalization deterioration when the ratio of seen classes reduces to a certain degree. 
%The `Global' baseline behaves as the best global-feature-based solution in most cases, but it still suffers severely from the generalizing issue and results in a lower accuracy than DeViSE for the 30/30 split of \textit{NTU-RGB+D 60}. 
Meanwhile, PURLS effectively mitigates this issue and consistently maintains its prediction preciseness.
% }

%
\subsection{Ablation Study}
\label{subsec: ablation}

% ablation study on universality
\begin{table}[]
\centering
\resizebox{\columnwidth}{!}{%
\begin{threeparttable}
\begin{tabular}{@{}cccccccccc@{}}
\toprule
\multirow{2}{*}{\textbf{Encoder}} & \multirow{2}{*}{\textbf{Descriptor}} & \multirow{2}{*}{\textbf{Model}} & \multicolumn{4}{c}{\textbf{NTU-RGBD 60 (Acc \%)}} \\
          &                &  & 55/5 & 48/12 & 40/20 & 30/30 \\ \midrule
%Shift-GCN & text-davinci-003 & 79.23     &  40.99     &  31.05     &  23.52     \\

% Global baseline
AA \cite{aa-gcn}   & GPT3 & Global & 62.79     &   28.09    &  25.66     &  13.86     \\
AA \cite{aa-gcn}   & GPT3 & PURLS & 76.75     &   32.39    &  31.00     &  21.86     \\
CTR \cite{ctr-gcn}   & GPT3 & Global & 65.16     &  34.56     & 26.12      & 15.92      \\
CTR \cite{ctr-gcn}   & GPT3 & PURLS & 79.97     &  39.42     & 32.26      & 24.59      \\
DG \cite{dg-gcn}    & GPT3 & Global  & 64.28     &  34.04     & 27.63      & 16.71      \\
DG \cite{dg-gcn}    & GPT3 & PURLS  & 80.41     &  41.06     & 33.77      & 25.12      \\
PoseC3D \cite{c3d}    & GPT3 & Global  & 63.45     &  35.71     & 27.88      & 20.66      \\
PoseC3D \cite{c3d}    & GPT3 & PURLS  & 81.14     &  41.60     & \textbf{34.47}      & \textbf{28.11}      \\
Shift & GPT3 & Global  & 64.69     &  35.46     &  27.15     &  16.29     \\
Shift & GPT3 & PURLS  & 79.23     &  40.99     &  31.05     &  23.52     \\
Shift & GPT3.5 & Global    & 66.49    & 38.01      &  26.31     &   17.35    \\
Shift & GPT3.5 & PURLS    &  79.17    & 40.98      &  30.07     &   {19.95}    \\Shift & GPT4 & Global             &  64.71    & 40.76      &  25.68     &  20.58    \\
Shift & GPT4 & PURLS             &  \textbf{81.53}    & \textbf{41.90}      &  {27.28}     &  {21.45}    \\\bottomrule
% PURLS
%\hline
\end{tabular}
\end{threeparttable}}
\caption{Ablation study on \textit{NTU-RGB+D 60} (\%) for examining the universality of PURLS by replacing the skeleton encoder backbone or the action descriptor. }
\label{table:universality}
\end{table}

% Ablation on partitioning strategy
\begin{table}[]
\centering
\resizebox{\columnwidth}{!}{%
\begin{tabular}{ccccccccc}
\toprule
                                          & \multicolumn{4}{c}{\textbf{NTU-RGBD 60 (Acc \%)} }           & \multicolumn{4}{c}{\textbf{NTU-RGBD 120 (Acc \%)}} \\
\multirow{-2}{*}{\textbf{\begin{tabular}[c]{@{}c@{}}Partitioning \\  Strategy\end{tabular}}} & 55/5                        & 48/12 & 40/20 & 30/30 & 110/10    & 96/24    & 80/40    & 60/60   \\ \midrule
Global (Original)                         & 64.69                       & 35.46 & 27.15 & 16.29 & 66.96     & 44.27    & 21.31    & 14.12   \\
Global (GPT-3)                            & 78.50 & 33.47 & 29.21 & 22.27 & 64.89     & 47.15    & 25.16    & 17.46   \\
Static                       & 76.46                       & 33.03 & 29.57 & 22.00    & 67.62     & 46.83    & 26.98    & 18.03   \\
Adaptive & \textbf{79.23} & \textbf{40.99} & \textbf{31.05} & \textbf{23.52} & \textbf{71.95} & \textbf{52.01} & \textbf{28.38} & \textbf{19.63} \\\bottomrule
\end{tabular}%
}
\caption{
% \textcolor{red}{
Ablation study (\%) on \textit{NTU-RGB+D 60} and \textit{NTU-RGB+D 120} for using different alignment learning with/without partitioning strategies, including direct global feature alignment to label or global description semantics, and PURLS with static/adaptive partitioning.}
% }
\label{table:ablation}
\end{table}

% Ablation on alpha, gb, bp & tp
\begin{table}[ht]
\centering
\resizebox{\columnwidth}{!}{%
{%
\begin{threeparttable}
\begin{tabular}{cllcccccccc}
\toprule
\multirow{2}{*}{$\bm{\alpha_i}$} &
  \multicolumn{1}{c}{\multirow{2}{*}{\textbf{BP}}} &
  \multicolumn{1}{c}{\multirow{2}{*}{\textbf{TI}}} &
  \multicolumn{4}{c}{\textbf{NTU-RGBD 60 (Acc \%)}} &
  \multicolumn{4}{c}{\textbf{NTU-RGBD 120 (Acc \%)}} \\ 
 &
  \multicolumn{1}{c}{} &
  \multicolumn{1}{c}{} &
  55/5 &
  48/12 &
  40/20 &
  30/30 &
  110/10 &
  96/24 &
  80/40 &
  60/60 \\ \midrule
-  &  &  & 78.50 & 33.47 & 29.21 & 22.27 & 64.89     & 47.15    & 25.16    & 17.46          \\ \midrule 
% Learnable &  &  & 66.15 & 33.03          & 22.38 & 16.50 & 40.55 & 33.56 & 16.49 & 10.84          \\ \midrule 
Average &\checkmark  &  & 76.68 & 37.80          & 30.92 & 22.20 & 68.11 & 30.93 & 24.36 & 18.67 \\ 
Learnable &\checkmark  &  & 76.32 & 37.62          & 29.06 & 21.91 & 71.73 & 40.92 & 23.49 & 19.13          \\ \midrule 
Average &  & \checkmark & 78.65 & 38.80 & 28.14 & 22.69 & 55.73 & 50.67 & 27.50 & 17.50          \\ 
Learnable &  & \checkmark & 77.70 & 40.69          & 28.84 & 22.46 & 71.26 & 46.13 & 24.43 & 18.57          \\ \midrule 
Average & \checkmark & \checkmark & 79.02 & 39.92          & 31.00 & 23.47 & \textbf{73.55} & 51.38 & 27.67 & 18.66          \\ 
Learnable & \checkmark
   & \checkmark   &
  \textbf{79.23} &
  \textbf{40.99} &
  \textbf{31.05} &
  \textbf{23.52} &
  71.95 &
  \textbf{52.01} &
  \textbf{28.38} &
  \textbf{19.63} \\ \bottomrule
\end{tabular}%
  \end{threeparttable}}
}
\caption{
% \textcolor{red}{
Ablation study (\%) on \textit{NTU-RGB+D 60} and \textit{NTU-RGB+D 120} for (1) using different $\alpha_i$ ($i\in[0, P+Z+1)$) to sum for $L_{train}$, (2) adding body-part-based (BP) alignment learning, (3) adding temporal-interval-based (TI) alignment learning. Note that when $L_{train}$ only contains global alignment learning (Row 1), $\alpha_i$ is not applicable.}
% }
\label{table:ablation2}
\end{table}

% \textcolor{red}{
We borrowed the setups in the \textit{NTU-RGB+D} series to conduct a careful ablation study on PURLS. We analyzed the method universality with auxiliary benefits from using description-based textual features and incorporating local semantic alignment. This includes the examination of four factors: (1) the universality among different skeleton backbones and description generators, (2) the disparity between using original labels and expanded descriptions for global feature alignment learning, (3) the efficiency of various partitioning strategies when sampling local visual concepts, (4) the respective influence of distilling spatially and temporally local knowledge for global prediction. 
% }

% \textcolor{red}{
\textbf{Universality:} \cref{table:universality} illustrates the detailed performance of PURLS on \textit{NTU-RGB+D 60} when it uses different skeleton encoders $g$ for visual feature extraction and GPT models for description generation. As a comparison, we also test the replacements on the `Global' baseline (\ie, aligning between the globally averaged skeleton features and label semantics) to verify the improvements brought by our method. For the skeleton backbone, we considered alternatives from several state-of-the-art extractors, including AA-GCN\cite{aa-gcn}, CTR-GCN\cite{ctr-gcn}, DG-STGCN\cite{dg-gcn}, and PoseC3D\cite{c3d}. In particular, PoseC3D is a unique backbone whose output format is a convoluted feature map from its 2-D heatmap input processed from an original skeleton sequence. In this situation, static partitioning is no longer compatible because it cannot pre-define which feature map pixels should belong to a specified body part. On the other hand, PURLS can still easily adapt itself to the new input structure by finding pixel-wise attention weights when generating a global/local visual representation. For the GPT descriptors, we attempted replacements based on model versions, iterating from GPT-3 to GPT-4. The results show that PURLS achieves an absolute advantage over `Global' across all examined settings, revealing that our solution steadily supports a better ZSL ability for most skeleton backbones and language models.
 % }

% \begin{MyColorPar}{red}
\textbf{Label Semantics vs. Description Semantics}: The first two rows of \cref{table:ablation} present the performance difference of only learning global feature alignment using label semantics or action description semantics. In most scenarios, description-based learning effectively boosts the results with richer semantic correlations across seen and unseen classes.

\textbf{Partitioning Strategy}: The following two rows of \cref{table:ablation} demonstrate the performance of PURLS using either static or adaptive partitioning. Using static partitioning shows unstable performance improvement compared to only learning global alignment from action descriptions. This is natural since the manual joint assignment may not capture all meaningful local movements in a given action but can also introduce noises. On the other hand, adaptive partitioning can effectively ameliorate these defects. 

\textbf{Local Semantics \& Aggregation}: In \cref{table:ablation2}, we further examine the concrete improvement brought by distilling the knowledge of body-part-based and temporal-interval-based local movements from adaptive partitioning. For the aggregation method of different alignment losses, we provided two options to sum the contrastive loss $\mathcal{L}(V_i, F_i)$ for each global/local representation $i$, including either averaging the weight of each term or applying a learnable weight (see \cref{eq:loss}). According to the results on Row 2-7, we find that the extra alignment losses from spatial and temporal dimensions can bring various prediction accuracy increases. This reveals that the model managed to extract local transferrable knowledge that improves the generalization of predictions for unseen classes. By adaptively distilling action-relevant knowledge from all possible global/local scales in a balanced manner, PURLS can achieve a robust recognition enhancement.
% \end{MyColorPar}

%% file: sec/5_conclusion.tex
\section{Conclusion}

We have introduced a novel framework, PURLS, that globally and locally aligns language and skeleton feature representations. We implement this by leveraging label semantic enrichment with large language models, as well as adaptive node feature partitioning on the skeleton structure. This enables PURLS to transfer various visual knowledge from seen classes to unseen classes at different scales. Experimental results demonstrate that PURLS achieves state-of-the-art performance not only in the existing ZSL setups on the \textit{NTU-RGB+D} series, but also in the more challenging setups of a customized dataset \textit{Kinetics-skeleton 200}. Furthermore, 
% \textcolor{red}{
PURLS shows powerful generality on different backbones and 
% }
effectively mitigates the generalization drops when the pre-training on seen classes is severely limited. It holds promise for future related cross-modal tasks by providing flexible feature alignment between natural language and joint-based motion data.

%% file: sec/6_acknowledgement.tex
\section{Acknowledgement}
% This research was undertaken using the LIEF HPC-GPGPU Facility hosted at the University of Melbourne. This Facility was established with the assistance of LIEF Grant LE170100200. MG was supported by ARC DE210101624.
This research was supported by The University of Melbourne’s Research Computing Services and the Petascale Campus Initiative.

%% file: main.bbl
\begin{thebibliography}{49}
\providecommand{\natexlab}[1]{#1}
\providecommand{\url}[1]{\texttt{#1}}
\expandafter\ifx\csname urlstyle\endcsname\relax
  \providecommand{\doi}[1]{doi: #1}\else
  \providecommand{\doi}{doi: \begingroup \urlstyle{rm}\Url}\fi

\bibitem[Armeni et~al.(2016)Armeni, Sener, Zamir, Jiang, Brilakis, Fischer, and Savarese]{vr1}
Iro Armeni, Ozan Sener, Amir~R. Zamir, Helen Jiang, Ioannis Brilakis, Martin Fischer, and Silvio Savarese.
\newblock 3d semantic parsing of large-scale indoor spaces.
\newblock In \emph{Proceedings of the IEEE Conference on Computer Vision and Pattern Recognition (CVPR)}, 2016.

\bibitem[Brown et~al.(2020)Brown, Mann, Ryder, Subbiah, Kaplan, Dhariwal, Neelakantan, Shyam, Sastry, Askell, Agarwal, Herbert-Voss, Krueger, Henighan, Child, Ramesh, Ziegler, Wu, Winter, Hesse, Chen, Sigler, Litwin, Gray, Chess, Clark, Berner, McCandlish, Radford, Sutskever, and Amodei]{gpt3}
Tom~B. Brown, Benjamin Mann, Nick Ryder, Melanie Subbiah, Jared Kaplan, Prafulla Dhariwal, Arvind Neelakantan, Pranav Shyam, Girish Sastry, Amanda Askell, Sandhini Agarwal, Ariel Herbert-Voss, Gretchen Krueger, Tom Henighan, Rewon Child, Aditya Ramesh, Daniel~M. Ziegler, Jeffrey Wu, Clemens Winter, Christopher Hesse, Mark Chen, Eric Sigler, Mateusz Litwin, Scott Gray, Benjamin Chess, Jack Clark, Christopher Berner, Sam McCandlish, Alec Radford, Ilya Sutskever, and Dario Amodei.
\newblock Language models are few-shot learners, 2020.

\bibitem[Bucher et~al.(2016)Bucher, Herbin, and Jurie]{metric}
Maxime Bucher, St{\'{e}}phane Herbin, and Fr{\'{e}}d{\'{e}}ric Jurie.
\newblock Improving semantic embedding consistency by metric learning for zero-shot classification.
\newblock \emph{CoRR}, abs/1607.08085, 2016.

\bibitem[Cadena et~al.(2016)Cadena, Dick, and Reid]{robotics2}
Cesar Cadena, Anthony Dick, and Ian Reid.
\newblock Multi-modal auto-encoders as joint estimators for robotics scene understanding.
\newblock 2016.

\bibitem[Cao et~al.(2021)Cao, Liu, Huang, Sheng, and Ju]{deep-skel}
Yi Cao, Chen Liu, Zilong Huang, Yongjian Sheng, and Yongjian Ju.
\newblock Skeleton-based action recognition with temporal action graph and temporal adaptive graph convolution structure.
\newblock 80\penalty0 (19):\penalty0 29139--29162, 2021.

\bibitem[{Cao} et~al.(2019){Cao}, {Hidalgo Martinez}, {Simon}, {Wei}, and {Sheikh}]{openpose}
Z. {Cao}, G. {Hidalgo Martinez}, T. {Simon}, S. {Wei}, and Y.~A. {Sheikh}.
\newblock Openpose: Realtime multi-person 2d pose estimation using part affinity fields.
\newblock \emph{IEEE Transactions on Pattern Analysis and Machine Intelligence}, 2019.

\bibitem[Chen et~al.(2015)Chen, Jafari, and Kehtarnavaz]{utdmhad}
Chen Chen, Roozbeh Jafari, and Nasser Kehtarnavaz.
\newblock Utd-mhad: A multimodal dataset for human action recognition utilizing a depth camera and a wearable inertial sensor.
\newblock In \emph{2015 IEEE International Conference on Image Processing (ICIP)}, pages 168--172, 2015.

\bibitem[Chen et~al.(2022)Chen, Hong, Xie, Yang, Peng, Wang, Zhao, and You]{msdn}
Shiming Chen, Ziming Hong, Guo-Sen Xie, Wenhan Yang, Qinmu Peng, Kai Wang, Jian Zhao, and Xinge You.
\newblock Msdn: Mutually semantic distillation network for zero-shot learning, 2022.

\bibitem[Chen et~al.(2021{\natexlab{a}})Chen, Zhang, Yuan, Li, Deng, and Hu]{ctr-gcn}
Yuxin Chen, Ziqi Zhang, Chunfeng Yuan, Bing Li, Ying Deng, and Weiming Hu.
\newblock Channel-wise topology refinement graph convolution for skeleton-based action recognition.
\newblock In \emph{Proceedings of the IEEE/CVF International Conference on Computer Vision}, pages 13359--13368, 2021{\natexlab{a}}.

\bibitem[Chen et~al.(2021{\natexlab{b}})Chen, Luo, Qiu, Wang, Huang, Li, and Zhang]{semantic-disentangle}
Zhi Chen, Yadan Luo, Ruihong Qiu, Sen Wang, Zi Huang, Jingjing Li, and Zheng Zhang.
\newblock Semantics disentangling for generalized zero-shot learning, 2021{\natexlab{b}}.

\bibitem[Cheng et~al.(2020)Cheng, Zhang, He, Chen, Cheng, and Lu]{shift-gcn}
Ke Cheng, Yifan Zhang, Xiangyu He, Weihan Chen, Jian Cheng, and Hanqing Lu.
\newblock Skeleton-based action recognition with shift graph convolutional network.
\newblock In \emph{2020 IEEE/CVF Conference on Computer Vision and Pattern Recognition (CVPR)}, pages 180--189, 2020.

\bibitem[Dinu et~al.(2015)Dinu, Lazaridou, and Baroni]{emb3}
Georgiana Dinu, Angeliki Lazaridou, and Marco Baroni.
\newblock Improving zero-shot learning by mitigating the hubness problem, 2015.

\bibitem[Duan et~al.(2021)Duan, Zhao, Chen, Lin, and Dai]{c3d}
Haodong Duan, Yue Zhao, Kai Chen, Dahua Lin, and Bo Dai.
\newblock Revisiting skeleton-based action recognition.
\newblock \emph{arXiv preprint arXiv:2104.13586}, 2021.

\bibitem[Frome et~al.(2013)Frome, Corrado, Shlens, Bengio, Dean, Ranzato, and Mikolov]{devise}
Andrea Frome, Greg~S Corrado, Jon Shlens, Samy Bengio, Jeff Dean, Marc\textquotesingle~Aurelio Ranzato, and Tomas Mikolov.
\newblock Devise: A deep visual-semantic embedding model.
\newblock In \emph{Advances in Neural Information Processing Systems}. Curran Associates, Inc., 2013.

\bibitem[Gupta et~al.(2021)Gupta, Sharma, and Sarvadevabhatla]{synse-zsl}
Pranay Gupta, Divyanshu Sharma, and Ravi~Kiran Sarvadevabhatla.
\newblock Syntactically guided generative embeddings for zero-shot skeleton action recognition.
\newblock In \emph{2021 IEEE International Conference on Image Processing (ICIP)}, pages 439--443, 2021.

\bibitem[Kay et~al.(2017)Kay, Carreira, Simonyan, Zhang, Hillier, Vijayanarasimhan, Viola, Green, Back, Natsev, Suleyman, and Zisserman]{kinetic}
Will Kay, Joao Carreira, Karen Simonyan, Brian Zhang, Chloe Hillier, Sudheendra Vijayanarasimhan, Fabio Viola, Tim Green, Trevor Back, Paul Natsev, Mustafa Suleyman, and Andrew Zisserman.
\newblock The kinetics human action video dataset, 2017.

\bibitem[Kodirov et~al.(2017)Kodirov, Xiang, and Gong]{aezsl}
Elyor Kodirov, Tao Xiang, and Shaogang Gong.
\newblock Semantic autoencoder for zero-shot learning, 2017.

\bibitem[Lazaridou et~al.(2015)Lazaridou, Dinu, and Baroni]{emb2}
Angeliki Lazaridou, Georgiana Dinu, and Marco Baroni.
\newblock Hubness and pollution: Delving into cross-space mapping for zero-shot learning.
\newblock pages 270--280, 2015.

\bibitem[Li et~al.(2019)Li, Chen, Chen, Zhang, Wang, and Tian]{deep-skel3}
Maosen Li, Siheng Chen, Xu Chen, Ya Zhang, Yanfeng Wang, and Qi Tian.
\newblock Actional-structural graph convolutional networks for skeleton-based action recognition.
\newblock In \emph{CVPR}, pages 3595--3603. Computer Vision Foundation / {IEEE}, 2019.

\bibitem[Li et~al.(2022)Li, Yu, Meng, Caine, Ngiam, Peng, Shen, Wu, Lu, Zhou, Le, Yuille, and Tan]{autodrive1}
Yingwei Li, Adams~Wei Yu, Tianjian Meng, Ben Caine, Jiquan Ngiam, Daiyi Peng, Junyang Shen, Bo Wu, Yifeng Lu, Denny Zhou, Quoc~V. Le, Alan Yuille, and Mingxing Tan.
\newblock Deepfusion: Lidar-camera deep fusion for multi-modal 3d object detection, 2022.

\bibitem[Liu et~al.(2019)Liu, Shahroudy, Perez, Wang, Duan, and Kot]{ntu120}
Jun Liu, Amir Shahroudy, Mauricio Perez, Gang Wang, Ling-Yu Duan, and Alex~C Kot.
\newblock Ntu rgb+ d 120: A large-scale benchmark for 3d human activity understanding.
\newblock \emph{IEEE transactions on pattern analysis and machine intelligence}, 42\penalty0 (10):\penalty0 2684--2701, 2019.

\bibitem[Nie and Liu(2021)]{deep-skel2}
Qiang Nie and Yunhui Liu.
\newblock View transfer on human skeleton pose: Automatically disentangle the view-variant and view-invariant information for pose representation learning.
\newblock \emph{IJCV}, 129\penalty0 (1):\penalty0 1--22, 2021.

\bibitem[Parsa et~al.(2020)Parsa, Narayanan, and Dariush]{pyramid}
Behnoosh Parsa, Athma Narayanan, and Behzad Dariush.
\newblock Spatio-temporal pyramid graph convolutions for human action recognition and postural assessment.
\newblock pages 1069--1079. {IEEE}, 2020.

\bibitem[Pham et~al.(2022)Pham, Nguyen, Nguyen, Nguyen, Nguyen, Pham, Tran, Le, and Vu]{app3}
Quang{-}Tien Pham, Duc{-}Anh Nguyen, Tien{-}Thanh Nguyen, Thanh~Nam Nguyen, Duy{-}Tung Nguyen, Dinh{-}Tan Pham, Thanh{-}Hai Tran, Thi{-}Lan Le, and Hai Vu.
\newblock A study on skeleton-based action recognition and its application to physical exercise recognition.
\newblock In \emph{The 11th International Symposium on Information and Communication Technology, SoICT 2022, Hanoi, Vietnam, December 1-3, 2022}, pages 239--246. {ACM}, 2022.

\bibitem[Qian et~al.(2022)Qian, Yu, Liu, and Hauptmann]{temp_align}
Yijun Qian, Lijun Yu, Wenhe Liu, and Alexander~G Hauptmann.
\newblock Rethinking zero-shot action recognition: Learning from latent atomic actions.
\newblock In \emph{European Conference on Computer Vision}, pages 104--120. Springer, 2022.

\bibitem[Radford et~al.(2021)Radford, Kim, Hallacy, Ramesh, Goh, Agarwal, Sastry, Askell, Mishkin, Clark, Krueger, and Sutskever]{clip}
Alec Radford, Jong~Wook Kim, Chris Hallacy, Aditya Ramesh, Gabriel Goh, Sandhini Agarwal, Girish Sastry, Amanda Askell, Pamela Mishkin, Jack Clark, Gretchen Krueger, and Ilya Sutskever.
\newblock Learning transferable visual models from natural language supervision, 2021.

\bibitem[Rahmani et~al.(2015)Rahmani, Mahmood, Huynh, and Mian]{uwa3dii}
Hossein Rahmani, Arif Mahmood, Du Huynh, and Ajmal Mian.
\newblock Histogram of oriented principal components for cross-view action recognition, 2015.

\bibitem[Rashmi and Guddeti(2020)]{app1}
M Rashmi and Ram Mohana~Reddy Guddeti.
\newblock Skeleton based human action recognition for smart city application using deep learning.
\newblock In \emph{2020 international conference on communication systems \& networks (COMSNETS)}, pages 756--761. IEEE, 2020.

\bibitem[Schonfeld et~al.(2019)Schonfeld, Ebrahimi, Sinha, Darrell, and Akata]{cada-vae}
E. Schonfeld, S. Ebrahimi, S. Sinha, T. Darrell, and Z. Akata.
\newblock Generalized zero- and few-shot learning via aligned variational autoencoders.
\newblock In \emph{2019 IEEE/CVF Conference on Computer Vision and Pattern Recognition (CVPR)}, pages 8239--8247, Los Alamitos, CA, USA, 2019. IEEE Computer Society.

\bibitem[Shahroudy et~al.(2016)Shahroudy, Liu, Ng, and Wang]{ntu60}
Amir Shahroudy, Jun Liu, Tian-Tsong Ng, and Gang Wang.
\newblock Ntu rgb+d: A large scale dataset for 3d human activity analysis.
\newblock In \emph{2016 IEEE Conference on Computer Vision and Pattern Recognition (CVPR)}, pages 1010--1019, 2016.

\bibitem[Shi et~al.(2019{\natexlab{a}})Shi, Zhang, Cheng, and Lu]{2s-gcn}
Lei Shi, Yifan Zhang, Jian Cheng, and Hanqing Lu.
\newblock Two-stream adaptive graph convolutional networks for skeleton-based action recognition.
\newblock In \emph{CVPR}, pages 12026--12035. Computer Vision Foundation / {IEEE}, 2019{\natexlab{a}}.

\bibitem[Shi et~al.(2019{\natexlab{b}})Shi, Zhang, Cheng, and Lu]{dg-gcn}
Lei Shi, Yifan Zhang, Jian Cheng, and Hanqing Lu.
\newblock Skeleton-based action recognition with directed graph neural networks.
\newblock In \emph{CVPR}, pages 7912--7921. Computer Vision Foundation / {IEEE}, 2019{\natexlab{b}}.

\bibitem[Shi et~al.(2020)Shi, Zhang, Cheng, and Lu]{aa-gcn}
Lei Shi, Yifan Zhang, Jian Cheng, and Hanqing Lu.
\newblock Skeleton-based action recognition with multi-stream adaptive graph convolutional networks.
\newblock \emph{IEEE Transactions on Image Processing}, 29:\penalty0 9532--9545, 2020.

\bibitem[Sun et~al.(2022)Sun, Ke, Rahmani, Bennamoun, Wang, and Liu]{har-survey}
Zehua Sun, Qiuhong Ke, Hossein Rahmani, Mohammed Bennamoun, Gang Wang, and Jun Liu.
\newblock Human action recognition from various data modalities: A review.
\newblock \emph{{IEEE} Transactions on Pattern Analysis and Machine Intelligence}, pages 1--20, 2022.

\bibitem[Tran et~al.(2018)Tran, Wang, Torresani, Ray, LeCun, and Paluri]{r2+1d}
Du Tran, Heng Wang, Lorenzo Torresani, Jamie Ray, Yann LeCun, and Manohar Paluri.
\newblock A closer look at spatiotemporal convolutions for action recognition.
\newblock In \emph{CVPR}, pages 6450--6459. Computer Vision Foundation / {IEEE} Computer Society, 2018.

\bibitem[Tsai et~al.(2017)Tsai, Huang, and Salakhutdinov]{revise}
Yao-Hung~Hubert Tsai, Liang-Kang Huang, and Ruslan Salakhutdinov.
\newblock Learning robust visual-semantic embeddings, 2017.

\bibitem[Vu et~al.(2022)Vu, Kim, Luu, Nguyen, and Yoo]{vr2}
Thang Vu, Kookhoi Kim, Tung~M. Luu, Xuan~Thanh Nguyen, and Chang~D. Yoo.
\newblock Softgroup for 3d instance segmentation on point clouds, 2022.

\bibitem[wang et~al.(2014)wang, Nie, Xia, Wu, and Zhu]{nwucla}
Jiang wang, Xiaohan Nie, Yin Xia, Ying Wu, and Song-Chun Zhu.
\newblock Cross-view action modeling, learning and recognition, 2014.

\bibitem[Wojek et~al.(2011)Wojek, Walk, Roth, and Schiele]{robotics1}
Christian Wojek, Stefan Walk, Stefan Roth, and Bernt Schiele.
\newblock Monocular 3d scene understanding with explicit occlusion reasoning.
\newblock pages 1993--2000, 2011.

\bibitem[Wray et~al.(2019)Wray, Csurka, Larlus, and Damen]{jpose}
Michael Wray, Gabriela Csurka, Diane Larlus, and Dima Damen.
\newblock Fine-grained action retrieval through multiple parts-of-speech embeddings.
\newblock In \emph{2019 IEEE/CVF International Conference on Computer Vision (ICCV)}, pages 450--459, 2019.

\bibitem[Xiang et~al.(2023)Xiang, Li, Zhou, Wang, and Zhang]{gap}
Wangmeng Xiang, Chao Li, Yuxuan Zhou, Biao Wang, and Lei Zhang.
\newblock Generative action description prompts for skeleton-based action recognition.
\newblock In \emph{{IEEE/CVF} International Conference on Computer Vision, {ICCV} 2023, Paris, France, October 1-6, 2023}, pages 10242--10251. {IEEE}, 2023.

\bibitem[Xue et~al.(2023)Xue, Gao, Xing, Martín-Martín, Wu, Xiong, Xu, Niebles, and Savarese]{ulip}
Le Xue, Mingfei Gao, Chen Xing, Roberto Martín-Martín, Jiajun Wu, Caiming Xiong, Ran Xu, Juan~Carlos Niebles, and Silvio Savarese.
\newblock Ulip: Learning a unified representation of language, images, and point clouds for 3d understanding, 2023.

\bibitem[Yan et~al.(2018)Yan, Xiong, and Lin]{st-gcn}
Sijie Yan, Yuanjun Xiong, and Dahua Lin.
\newblock Spatial temporal graph convolutional networks for skeleton-based action recognition, 2018.

\bibitem[Yin et~al.(2021)Yin, Zhou, and Krahenbuhl]{autodrive2}
Tianwei Yin, Xingyi Zhou, and Philipp Krahenbuhl.
\newblock Center-based 3d object detection and tracking.
\newblock In \emph{Proceedings of the IEEE/CVF Conference on Computer Vision and Pattern Recognition (CVPR)}, pages 11784--11793, 2021.

\bibitem[Zhang et~al.(2019)Zhang, Xiang, and Gong]{emb1}
Li Zhang, Tao Xiang, and Shaogang Gong.
\newblock Learning a deep embedding model for zero-shot learning, 2019.

\bibitem[Zhang et~al.(2022)Zhang, Guo, Zhang, Li, Miao, Cui, Qiao, Gao, and Li]{point-clip}
Renrui Zhang, Ziyu Guo, Wei Zhang, Kunchang Li, Xupeng Miao, Bin Cui, Yu Qiao, Peng Gao, and Hongsheng Li.
\newblock Pointclip: Point cloud understanding by clip.
\newblock In \emph{Proceedings of the IEEE/CVF Conference on Computer Vision and Pattern Recognition (CVPR)}, pages 8552--8562, 2022.

\bibitem[Zhang et~al.(2021)Zhang, Tian, Wu, and Chen]{app2}
Yunkai Zhang, Yinghong Tian, Pingyi Wu, and Dongfan Chen.
\newblock Application of skeleton data and long short-term memory in action recognition of children with autism spectrum disorder.
\newblock \emph{Sensors}, 21\penalty0 (2):\penalty0 411, 2021.

\bibitem[Zhou et~al.(2023)Zhou, Qiang, Rao, Lin, Su, and Wang]{smie}
Yujie Zhou, Wenwen Qiang, Anyi Rao, Ning Lin, Bing Su, and Jiaqi Wang.
\newblock Zero-shot skeleton-based action recognition via mutual information estimation and maximization.
\newblock In \emph{Proceedings of the 31st {ACM} International Conference on Multimedia, {MM} 2023, Ottawa, ON, Canada, 29 October 2023- 3 November 2023}, pages 5302--5310. {ACM}, 2023.

\bibitem[Zhu et~al.(2023)Zhu, Jia, Chen, Guo, and Liu]{retrieval}
Cunjuan Zhu, Qi Jia, Wei Chen, Yanming Guo, and Yu Liu.
\newblock Deep learning for video-text retrieval: a review.
\newblock \emph{International Journal of Multimedia Information Retrieval}, 12\penalty0 (1):\penalty0 3, 2023.

\end{thebibliography}
